\documentclass[]{amsart}

\usepackage{wrapfig}
\usepackage{amsmath,amssymb, amsfonts, mathtools}

\usepackage{newtxtext,newtxmath}       
\usepackage{helvet}         
\usepackage{courier}        
\usepackage{graphicx}        
\usepackage{multicol}        
\usepackage[bottom]{footmisc}  
\usepackage{amsmath, amsfonts, mathtools}       
\usepackage{verbatim, booktabs}
\usepackage{bbm, bm}
\usepackage{multirow}
\usepackage{csvsimple}
\usepackage{longtable}
\usepackage[capitalise,noabbrev]{cleveref}

\usepackage{tikz}
\usepackage{pgfplots,pgfplotstable}
\pgfplotsset{compat=1.17}
\usepackage{subcaption}
\usepackage{enumitem}
\usepackage{epstopdf}
\usepackage{varwidth}
\usepackage{color}

\usepackage{tabularx}
\usepackage[export]{adjustbox}

\usepackage[utf8]{inputenc} 

\newcommand{\R}{\mathbb{R}}

\newcommand{\N}{\mathbb{N}}

\DeclareMathOperator*{\argmax}{argmax}

\newcommand{\given}{\,|\,}

\theoremstyle{plain}

\theoremstyle{definition}
\newtheorem*{defn*}{Definition}

\ifpdf
\DeclareGraphicsExtensions{.eps,.pdf,.png,.jpg}
\else
\DeclareGraphicsExtensions{.eps}
\fi

\newcommand{\rev}[1]{{\color{black}{#1}}}
\newcommand{\new}[1]{{\color{red}{#1}}}

\newcommand*{\escape}[1]{\texttt{\textbackslash#1}}

\makeatletter
\def\@setthanks{\vspace{-\baselineskip}\def\thanks##1{\@par##1\@addpunct.}\thankses}
\makeatother

\begin{document}

\title{Topological Interpretability for Deep-Learning}

\author{Adam Spannaus$^1$}
    \email[Corresponding author]{spannausat@ornl.gov}
\thanks{This manuscript has been authored by UT-Battelle, LLC under Contract No. DE-AC05-00OR22725 
	with the U.S. Department of Energy. The United States Government retains and the publisher, 
	by accepting the article for publication, acknowledges that the United States Government 
	retains a non-exclusive, paid-up, irrevocable,world-wide license to publish or reproduce 
	the published form of this manuscript, or allow others to do so, for United States 
	Government purposes. The Department of Energy will provide public access to these 
	results of federally sponsored research in accordance with the DOE Public 
	Access Plan (http://energy.gov/downloads/doe-public-access-plan).}
\address[1]{Oak Ridge National Laboratory, Oak Ridge, TN 37830}
\author{Heidi A. Hanson$^1$}
\author{Lynne Penberthy$^2$}
\address[2]{National Cancer Institute, Bethesda, MD, USA}
\author{Georgia Tourassi$^1$}
\renewcommand{\shortauthors}{Spannaus et al.}

\begin{abstract}
	With the growing adoption of AI-based systems across  
	everyday life, the need to understand their
	decision-making mechanisms is correspondingly increasing. 
	The level at which we can trust the statistical inferences
	made from AI-based decision systems is an increasing concern, 
	especially in high-risk systems
	such as criminal justice or medical diagnosis, where incorrect inferences may 
	have tragic consequences. 
	Despite their successes in providing solutions to problems involving real-world data, 
	deep learning (DL) models cannot quantify the certainty of 
	their predictions. These models are frequently quite confident, 
	even when their solutions are incorrect. 

	This work presents a method to infer prominent 
	features in two DL classification models trained on
	clinical and non-clinical text by employing 
	techniques from topological and geometric data analysis.
	We create a graph of a model's 
	feature space and cluster the inputs into the graph's vertices by the similarity of
	features and prediction statistics. We then extract subgraphs 
	demonstrating high-predictive accuracy for a given label. These subgraphs contain a 
	wealth of information about features that the DL model has recognized as 
	relevant to its decisions. 
	We infer these features for a given label using a distance metric between
	probability measures, and demonstrate
	the stability of our method compared to the LIME and SHAP interpretability methods.
	This work establishes that we may gain insights into the decision mechanism
	of a DL model. This method allows us to ascertain if the model is making its decisions
	based on information germane to the problem or identifies extraneous patterns within the 
	data.
\end{abstract}
\keywords{Topological Data Analysis, Deep Learning, Interpretability}

\maketitle

\section{Introduction}
	Understanding the process through which any machine learning (ML) algorithm makes a decision is crucial for 
	high-risk domains: this is especially true in both criminal justice and healthcare. Whether humans either
	make a decision with 
	the assistance of an algorithmic process or delegate the entirety of the decision-making process, the need for
	insight is paramount. Indeed, the desiderata for assurance and reliability across AI was identified as a 
	grand challenge for establishing 
	foundational scientific AI advancements~\cite{ai4sci-doe}. Due in part to the black-box nature of many 
	AI-based algorithms~\cite{begoli2019need} the field of machine-assisted medical decision-making 
	has been slow to adopt this paradigm. Even when their solutions turn out to be incorrect, AI-based models 
	provide high confidence levels in their predictions~\cite{guo2017calibration}.
	The medical professional's hesitancy to bring 
	machine assisted decision-making into the clinical setting is understandable;
	typical deep learning (DL) models cannot either reliably quantify the confidence
	in their predictions, nor provide an explanation for any particular decision.
 
 	From a given set of inputs, medical professionals must
 	be able to trust the decisions of the DL model and understand why
 	it made a specific prediction to reliably use the predictions in their practice.
	However, as noted in~\cite{ribeiro2016should},
 	trusting the predictions of a model is not sufficient; one must have trust in the 
 	model as well. The idea of trusting predictions is straightforward, it answers the question 
	of whether or not
 	the end-users have sufficient confidence in a prediction to take 
 	action. 
	Trust in a model is more subtle; to engender trust, one must 
	demonstrate that a prediction is made for the right reasons.

    To increase the assurance in an AI model's predictions, 
    we propose an interpretability
	method that is based on the topological and geometric properties of the model's 
	feature space. 
	We distinguish salient features in the input data that both inform 
	the classification of each input, and identify clusters of inputs with similar
	features and classification performance.
	Moreover, as defined by~\cite{laugel2019dangers}, our proposed methodology is \emph{justified}. 
	The authors of~\cite{laugel2019dangers} argue that explanations from any interpretability 
	method must to be tied to the training data set, ensuring that the explanations
	are not the result of a modeling artifact, but rather an inherent feature of the data.
	Our proposed methodology includes training data in our topological construction of the 
	feature space, and both global and local explanations are given with respect to 
	the training data.
    
    We create a topological representation of an AI model's feature
    space through the Mapper algorithm~\cite{singh2007topological}.
    Our construction clusters inputs by their ground truth class into regions
    of high and low predictive accuracy. We then 
	engage in an unsupervised semantic analysis of the inputs
    in these clusters to better understand the trained model's decision mechanism.
    This analysis reveals important words to 
    our classification model both for inputs it classifies correctly, and those
    it predicts incorrectly. As a direct result of this analysis, we
    obtain insight into similarities between correctly classified data points,
    and why others are incorrectly classified.
	
	\subsection{\rev{Related Works}}
	
	\rev{To create explanations from a trained model, one approach is to study samples
	from the training set. Another technique is to use a more interpretable model, 
	e.g.,~\cite{breiman2017classification,agarwal2021neural}, or by
	tracing the first-order gradients back to the input data
	to identify feature importance~\cite{simonyan2013deep,shrikumar2017learning}.
	
	Other methods for post-hoc interpretability, such as LIME~\cite{ribeiro2016should}
	or SHAP~\cite{lundberg2017unified}, perturb the original data, then 
	sample from these perturbations to make their inferences.
	These methods then either fit a proxy model which is more interpretable~\cite{ribeiro2016should},
	or rank the feature importance which inform a particular classification~\cite{lundberg2017unified}.
	For similar inputs however, these methods can provide very different results.
	As shown in~\cite{alvarez2018robustness}, the explanations provided by LIME 
	on a collection of UCI datasets demonstrated both instability and high variance.   
	Moreover, these methods
	infer counter-factual explanations and are easily confused~\cite{slack2020fooling}. 
	Counter-factual explanations are a minimal perturbation necessary 
	to change the prediction of the proxy model, and may not generate explanations 
	that are connected to the ground truth~\cite{rudin2018please}.  
	Consequently, to establish both meaningful interpretations and assurances in the decisions 
	from an AI model, perturbation and sampling based approaches are insufficient. 
	
	The works~\cite{rathore2021topoact,rathore2023topobert} use methods
	from TDA to explain deep learning methods, but extract the relevant 
	features in a different manner than the method proposed herein.
	For an in-depth treatment on the topic of explainable AI, we refer the
	interested reader to the works of~\cite{molnar2020interpretable,pml2Book}
	and references therein.}
		
	\subsection{\rev{Contributions}}
	
	Here we present an interpretability method for any AI model that
	yields a topologically correct construction of the model's feature space.
	Our method yields both global and local
	views of the features informing a decision. For the global view, we 
	extract features associated with samples classified with high predictive accuracy
	and create a low-dimensional representation of our model's feature space.
	We provide a method for inferring the features related to the classifying
	any input to the model. 
	\rev{We demonstrate the stability of our method as compared with LIME~\cite{ribeiro2016should}
	and SHAP~\cite{lundberg2017unified} and present results on two datasets.}
	We present results on two different datasets and trained models.
	First, we consider a multi-task convolutional neural network (MTCNN) taking
	cancer pathology reports as input and predicting cancer phenotype~\cite{alawad2020automatic}.
	Secondly, we trained a convolutional neural network (CNN) on 
	the publicly available 20newsgroups dataset~\cite{20News}.
	Through our analysis of these datasets, we see that our method is able to identify
	relevant features in the training data which inform the trained model's 
	decision mechanism.
	
	The outline of this paper is as follows. In~\cref{sect:MM} we give a high-level descrption
	of TDA and a description of our method. Numerical
	results on both models 
	are presented in~\cref{sect:results} and we conclude with discussion in~\cref{sect:discussion}.

\section{Materials and Methods}\label{sect:MM}

	
	\subsection{Topological Data Analysis}
   
	Topological Data Analysis (TDA) is a field of applied mathematics which extracts
	valuable information
	about connections within large and high-dimensional datasets by applying abstract 
	ideas and techniques from topology. Three 
	key ideas that differentiate topology from other geometric methods for data analysis are:
	(a) \emph{coordinate invariance}; (b) \emph{deformation invariance}; and 
	(c) \emph{compressed representation}. 
	
	The notion of coordinate invariance arises naturally from one of the primary building blocks of topology,
	the concept of a metric space. A metric space is a set
	of points equipped with a notion of distance between any pair of points.
	Traditional
	methods of data analysis such as Principal Component Analysis (PCA) or multidimensional scaling
	embeds the data into Euclidean space, and thus swaps any data-centric notion of
	similarity for the traditional Euclidean metric. If the
	data is indeed non-Euclidean, this embedding distorts relevant features.
	Topological methods 
	are insensitive to the choice of distance metric and 
	measures of closeness between a point and a subset within the data-space.
	
	The property
	of deformation invariance is central to any discussion of topological methods, as this
	approach allows for 
	shapes to be stretched, contracted, and/or twisted in such a way that 
	no new holes are created, nor are existing holes closed. Any continuous transformation
	preserves the topological features present within the data.
	
	Lastly, the idea of compressed representation is best illustrated through an
	example. Consider a circle, which is composed of an infinite number of points.
	Topologically speaking, its essential feature is that the set of points forms a closed loop.
	Now if we instead consider a hexagon, it too forms a closed loop and is
	topologically equivalent to the circle. Although geometric curvature information is lost
	by adopting this point of view, such trade-offs are common throughout computational mathematics. 
	Indeed, viewing the circle
	as an hexagon, we may now use it in computations as the hexagon requires six points to encode, 
	as opposed to an infinite number for the circle.
	
	We apply techniques from the mathematical study of shape to extract 
	fundamental information about the intrinsic relationships present within our data. 
	Relying on the topological notion of `closeness', we may extract insights from our 
	data that are 
	unavailable through rigid geometric notions of shape and distance. The end result of our analysis
	is a low dimensional representation of our data that is amenable to both further mathematical
	analysis and visualization.
	
	\subsection{Mapper Graph}

   	We will now discuss the Mapper graph, the primary tool through which we create a topologically consistent
   	construction of an DL model's feature space.
   	\subsubsection{Mapper Graph}
   	The Mapper graph was proposed as an algorithmic
	method to study high-dimensional data via simplicial complexes through chosen 
	filter function(s)~\cite{singh2007topological}. These functions are used to indicate properties of the data,
	such as a kernel density estimate, distance, or eccentricity.  
	This construction has found applications in such
	diverse fields as spinal cord injuries, breast cancer omics studies,
	and voting trends~\cite{nielson2015topological,lum2013extracting,nicolau2011topology}.
	The method has been used as a feature selection tool
	for identifying subpopulations within a larger group~\cite{singh2007topological,nicolau2011topology}, and
	recently the algorithm
	has been applied in an error analysis and activation functions of a DL 
	model~\cite{carlsson2020fibers,rathore2021topoact}. 
	
	The methodology creates a
	low-dimensional representation of high-dimensional data while preserving the
	original data's topology. Similar to PCA or a spectral embedding, the
	Mapper construction may be thought of as an unsupervised clustering algorithm.
	In both PCA or a spectral embedding though, the data is assumed to be in Euclidean
	space; the Mapper construction creates a graph, or higher dimensional simplicial complex, 
	yielding a combinatorial representation of the data without any assumptions 
	on the data space~\cite{carlsson2009topology}.
	This provides a scaffold
	for a topological construction by reducing the problem of approximating a shape 
	to a linear algebra problem. 
	This process is analogous to reconstructing a complex shape from a
	suitably chosen combination of points, lines, triangles, and their higher dimensional analogues.

	The primary idea of the Mapper construction is to summarize a dataset by creating a neighbor graph of
	the data. To this end, we create subsets of the filter function's domain,
	and apply a clustering method to these subsets. The resulting graph captures 
	clusters of inputs that exist solely in one subset, resulting in a vertex of the graph or
	those inputs appearing in multiple subsets, which creates an edge of the graph. 
	Through this construction, we are able to realize two types of neighbor
	relationships in a dataset: (\emph{i}) 0-dim information, those points that are assigned
	to a specific cluster, resulting in a vertex; and 
	(\emph{ii}) 1-dim information, those points that are assigned to multiple clusters which
	produce an edge on the graph, i.e., these data points contain some shared attributes. 
	Consequently, the ensuing graph construction is topologically 
	faithful to the original point cloud.
	Inputs having similar features will form a connected subgraph and be
	disconnected from dissimilar inputs, thus shape and neighbor relationships in
	the original data are preserved through this construction.
	This result is is non-trivial, we omit the technical details here; 
	see~\cite{singh2007topological,carlsson2009topology,carriere2018statistical,hatcher2002algebraic} and
	references contained therein.

	\subsubsection{Mapper Algorithm}
	
	We will now describe the process of creating a Mapper graph. The construction begins with 
   	point cloud $\mathbb{X}_N = \{X_i\}_{i=1}^N, X_i\in\R^d$ with $d\in\N$, for
   	a collection of $N$ points sampled from a metric space $\mathcal{X}$.
   	Let $f:\mathcal{X}\rightarrow\R$ be continuous, which 
   	is called a \emph{filter function}.
   	Through this function,
	one may show the equivalence of the Mapper reconstruction and
   	the underlying topological space in question by observing that if
	$X, X'\in\mathcal{X}$
	are contained in the same topological feature, then
   	vertices of a graph which are connected by an edge of
	the pre-image of $f^{-1}$
	are in the same topological feature.

	\begin{figure*}
		\centering
		\includegraphics[width=\textwidth]{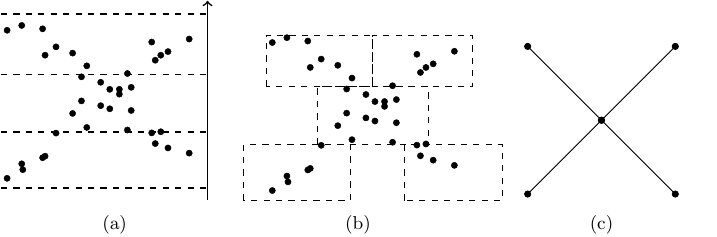}
		\caption{The Mapper algorithm proceeds as follows: In
			(a), observe a point cloud, filter function, chosen as the height function, and an open cover, denoted by the dashed lines. Fig. (b) we have covered the range of values in (a) by consecutive intervals.
			Clustering on the intervals in (b), and creating the \emph{nerve}. In fig. (c), we have the resulting 
			Mapper construction, demonstrating the compressed
			representation of the original data.}\label{fig:mapper}
	\end{figure*}

	Any collection of continuous functions that form an open cover of their image may be chosen as 
	the filter in a Mapper construction. The pre-images of this collection then form an open cover of the data space. 
	On this open cover of the data space, we apply a clustering algorithm,
	identifying the connected components of the data space. The clustering algorithm is iteratively 
	applied to subsets of the pre-image that are connected through the filter functions.
	Intuitively, one may think of this clustering process as repeated partial clusterings
	on the pre-image of the filter functions.
	This repeated clustering ensures similarity
	of features in each vertex of the Mapper graph. 
   	Employing different filter functions in a construction may highlight different aspects of a dataset. 
   	Indeed, as these functions separate the input data into
   	clusters that have different values under the filter function, one may create multiple
   	Mapper graphs from the same data, each with a different filter to emphasize
   	different aspects of the dataset. 
   	Visually, the algorithm proceeds as in~\cref{fig:mapper} and is written:
   	\begin{enumerate}
   		\item Choose $j$ filter functions such that $f_i:\mathcal{X}\rightarrow\R, 1\leq i\leq j$.
   		\item For each filter function, choose a number of intervals $S_i$ creating a 
   		cover $\mathbb{U}_i = \{U_s\}_{s=1}^{S_i}$ that covers $f_i(\mathbb{X}_N)$.
   		\item Apply clustering algorithm to the pre-image $f_i^{-1}(U_i)$, defining
   		a \emph{pullback cover} $\mathcal{C}$ on $\mathbb{X}_N$.
   		\item The Mapper graph $\mathcal{M}$ is the
   		\emph{nerve} of $\mathcal{C}$; each vertex of $\mathcal{M}$ 
   		corresponds to one $c_{s,k}\in\mathcal{C}$ and vertices $v_{s,k}, v_{s',k'}$
   		are connected iff $c_{s,k}\cap c_{s',k'}\neq\emptyset$.
   	\end{enumerate}

   	The choice of clustering algorithm is 
   	left to the user, as is defining the resolution (the number of intervals),
   	and gain (the percent overlap between intervals). We follow the guidelines
   	described in~\cite{carriere2018statistical} to set these two hyper-parameters.

   	\subsection{Mapper on an Deep Learning Model}
   	
	\subsubsection{Multi-task Deep Learning Model}
	
\begin{figure}
		\centering
		\includegraphics[width=\columnwidth]{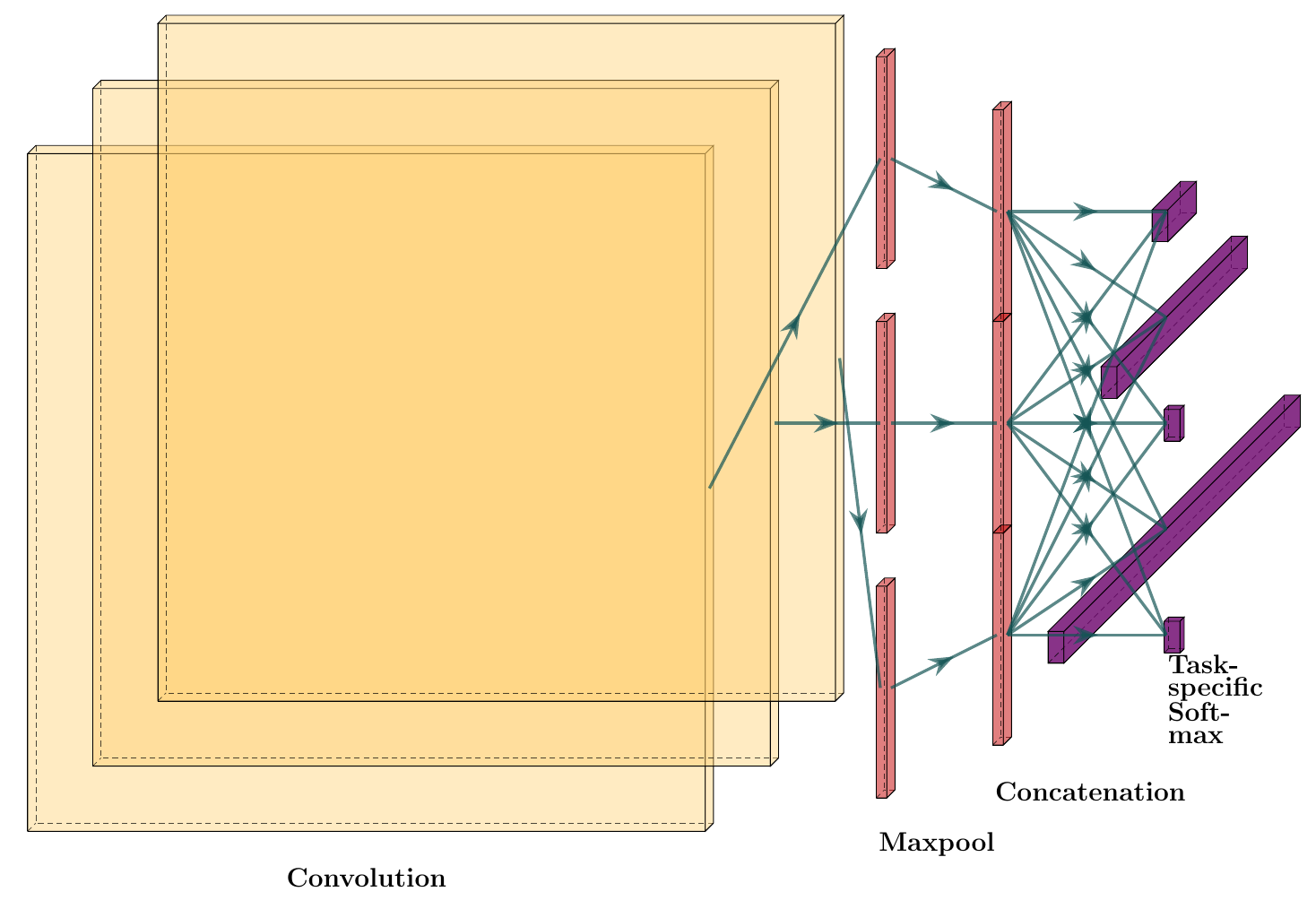}
		\caption{Our multi-task convolutional
			neural network architecture used for cancer pathology report information
			extraction. The network consists of three parallel convolutional filters, 
		 followed by a maxpooling and concatenation layer. To each maxpooling layer, we 
		 employ a  dropout rate of 50\%. After applying the dropout, we concatenate the remaining  
		 vectors and pass them to a task-specific softmax classification layer.}\label{fig:mtcnn}
	\end{figure}

	The MTCNN is a deep learning architecture for information
	extraction from cancer pathology reports~\cite{alawad2020automatic,GAO2019101726}. It
	takes as input a pathology report in which the words have been
	tokenized, i.e., mapped to a unique integer value, and outputs simultaneous predictions about five
	cancer phenotypes (tasks): primary site, subsite, histology, laterality, and grade.
	The parallel convolutional layers in the MTCNN model extracts features, which are concatenated into 
	a single feature vector (the document embedding), which is 
	subsequently passed to task-specific softmax classification layers.
	\Cref{fig:mtcnn} is a visual representation of our MTCNN architecture.
	
	\subsubsection{Mapper Graph of a Deep Learning Model}
	
	Crucial to gaining insight and interpretability of a model is the choice of the filter in the Mapper algorithm.
   	As previously noted, these functions enforce the separation
   	of data points with sufficiently different filter values into separate vertices in the
   	graph. By choosing the ground truth class of the input as one of the filters, we ensure
   	that the datapoints in any of the graph's vertices are homogeneous with respect to the ground truth.
   	Partitioning the input space in such a manner,  
   	we may investigate relevant features of our deep learning model by:
   	(\emph{i}) identifying clusters of inputs that are classified with high predictive accuracy
   	sharing the same label; or (\emph{ii}), identifying those features associated
   	with low-predictive accuracy for given class label.

	Our construction clusters input features by similarity, ground truth label, and predicted probability of 
	the ground truth label. Furthermore, these clusters contain documents with a similar 
	document embedding layer, i.e., the concatenation layer of our MTCNN model~(\cref{fig:mtcnn}), thus the individual
	features in the documents are similar.
	We seek to infer the pertinent features leading to both correct and incorrect classifications with
	a high probability, gaining insight into the DL model's decision mechanism.

   	Clustered by similarity of the document embedding layer in our MTCNN model, 
	the nodes of our Mapper graph may contain anywhere from a small handful to hundreds of documents. 
	To provide interpretable results for the end-user, we must
	identify the relevant features in these documents from regions of high 
   	predictive accuracy.

	\subsection{Geometric Analysis on a Probability Measure}\label{sect:dtm}

	In a classification problem, one seeks to infer the probability that a given collection of features
	belongs to $1$-of-$k$ classes.  
	The inverse question of interpretability may be posed in a similar manner,
	where the given data is a class
	label, and we seek a collection of features informing the particular classification.
	
	To infer the relevant features from the data informing a classification, we investigate
	the probability density
	\begin{equation}
		\pi(w\in \mathcal{V}\given\mathcal{H}),\label{eq:prob_mzr}
	\end{equation}
	where $\mathcal{H} = \{X\given p(y\given X)\geq\alpha, y\in\{1,\dots,k\},\alpha\in[0,1]\}$, is the set of 
	all inputs that have confidence scores from an AI model of at least $\alpha$. That is
	a model taking input $X\in\mathbb{X}_N$
	and outputs a confidence score $p(y\given X), y\in\{1,\dots,k\}$ for a $k$-class classification problem, 
	where $\hat{y} = \argmax_{y\in\{1,\dots,k\}} p(y\given X)$ is the predicted label of $X$.
	Lastly, we define
	$\mathcal{V} = \{(x, n)\given n\in\N, x = X_{i,*},X\in\mathcal{H}\}$ as the multiset of all entries of $X$, and
	$X_{i,*}$ is the $i^{th}$ row of $X$. We investigate 
	the support of~\cref{eq:prob_mzr} via the distance to measure (dtm) function of~\cite{Chazal2011}.

	We will present a summary of the theory necessary for our proposed methodology;
	for a thorough treatment, see~\cite{Chazal2011,chazal2016rates} and
	references therein. The dtm
	function was initially proposed as a technique for outlier detection
	from a noisy sample where the end goal was manifold reconstruction and inference of topological invariants.
	Formally, the dtm measures the distance from a Dirac mass to a 
	probability measure. We will first define the pseudo-distance $\delta_{\mu,m}$ which is required for the 
	dtm function~\cite{Chazal2011}, whose definition follows.
	
	\begin{defn*}
		Given any $0\leq m <1$ and a probability measure $\mu$ on $\R^d$, define the pseudo-distance $\delta_{\mu, m}$ as
		\begin{equation*}
			\delta_{\mu, m}: x\in\R^d\mapsto \inf\{r>0: \mu\left(\overline{B}(x, r)\right) > m\},
		\end{equation*}
	\end{defn*}
	\noindent where $\overline{B}(x, r)$ is the closed ball of radius $r$ centered at $x$. This is the probabilistic
	interpretation of a distance function that measures the distance from a point to a compact set. 

	We now define the distance to measure of~\cite{Chazal2011}.
	\begin{defn*}
		Assume that $\mu$ is a probability measure on $\R^n$ and $\hat{m}\in(0,1)$ is a parameter. Then
		the distance function to $\mu$ with parameter $\hat{m}$ is defined by:
		\begin{equation}
			\mathrm{d}^2_{\mu,\hat{m}}:\R^d\rightarrow \R^{+}, x\mapsto\frac1{\hat{m}}\int_0^{\hat{m}}\delta_{\mu,m}(x)^2\mathrm{d}m.
			\label{eq:dtm}
		\end{equation}
	\end{defn*}
	
	The function defined in~\cref{eq:dtm} retains properties of a traditional distance function, 
	but is less sensitive to noisy 
	outliers. Indeed, in the case that $m=0$ we recover the traditional Euclidean distance, 
	but when $m>0$, we see that the dtm
	will have greater distances than the traditional Euclidean distance, since some portion of mass $m$
	will be included in the closed ball centered at $x$.
	
	Through our Mapper construction, we have an topologically faithful reconstruction of $\mathcal{H}$ and
	want to study the distance from individual features to the probability measure defined in~\cref{eq:prob_mzr}.
	Our goal is to gain insight into the features that a DL model uses in its classification rules. 
	By examining the distances to the support of this probability measure, we may identify the key features 
	in the learned classification rules. 
	Intuitively, we may think of the dtm process as yielding an estimate of the density~\cite{Biau2011},
	where the density is over the distribution of words within nodes of our mapper graph.

	\subsection{Feature Extraction}
	
	While the output from the Mapper algorithm clusters documents with similar
	predictive accuracies into vertices; it does not yield relevant features
	present within the documents informing the clusters in the graph's vertices.
	As our Mapper construction partitions the output space by ground truth labels, we may 
	consider the connected subgraph for a specific ground truth label, and 
	infer the pertinent features informing the classifications, either
	correct or incorrect, for a specific ground truth label. 
	
	To infer the relevant features for a classification we investigate the structure
	of the conditional probability distribution defined in~\cref{eq:prob_mzr}.
	We compute the dtm values for the subset of the vocabulary present in the
	documents clustered in the graph's vertices having at least a predictive
	confidence level of $\alpha$ from the multiset of all words contained in the 
	documents with the same label. In practice this involves finding
	the $K-$nearest neighbors to each word in the label specific subset from
	the multiset of words contained in all documents with the same label.
	To then compute~\cref{eq:dtm} we use the discrete approximation
	\begin{equation}
		\hat{d^2}_{\mu, \hat{m}}(x) = \frac1{K}\sum_{i=1}^K\,\|x - X_i(x)\|^2,\label{eq:edtm}
	\end{equation}
	where $X_i(x)$ denotes an ordering of the data so that 
	$\|x - X_j(x)\|^2 \geq \|x - X_\ell(x)\|^2$ for $j<\ell$ and then 
	$\hat{d^2}_{\mu, \hat{m}}(x)$ is simply the average squared Euclidean distance to the 
	point $x$ of the $K$-nearest neighbors.
	Intuitively, we are interested in the word density in the graph vertices
	with a high predictive accuracy. Computing this density is not computationally
	tractable, due to the dimension of the word embedding vectors and
	the length of the documents. Previous works have noted the connection between 
	computing the dtm is related to the density estimation
	problem~\cite{Biau2011}, and we leave the problem of density estimation of the 
	word vectors for a future investigation.
	
	We choose $K$ dependent on the size of the vocabulary comprised of all words in the documents
	having the same label, so that $\hat{m} = [0.05, 0.25]$. This value controls
	the amount of mass contained in the ball centered at each point in the multiset, so 
	higher values will contain more points, and is dependent on the data itself. Some
	guidance in choosing this parameter is given in~\cite{Chazal2011}, but rules 
	for choosing this parameter optimally remains an open question.

\section{Results}\label{sect:results}

\subsection{Cancer Pathology Reports}
	
	The data for our information extraction task is comprised of cancer pathology reports 
	collected by the NCI
	Surveillance, Epidemiology, and End Results (SEER) program. These cancer pathology reports are collated
	and analyzed by the Louisiana Tumor Registry.
	
	\rev{The histologic characteristics of a specific tumor are described within a pathology
	report.	Trained pathologists describe the disease using highly technical and
	structured text, which characterizes the anatomy of a tumor as seen through a microscope. 
	These reports are the definitive source for a cancer diagnosis.
	Each tumor diagnosis is assigned a unique tumor ID,
	and a report is created, documenting these characteristics associated with that specific case.
	Additionally, each report contains metadata such as the patient identification number and date
	of the report in addition to clinical text describing the tumor and 
	the patient's clinical history.}
	The SEER registries abstract the information contained in these reports relevant to the 
	characteristics associated with primary cancer site, grade, histology, laterality, and behavior
	\rev{for each unique tumor as assessed by a trained pathologist.
	These reports are collected and aggregated }
	to monitor cancer incidence in population for cancer control and prevention, inform
	research, and advise on new and emerging research directions. 

\begin{figure}
		\centering
		\includegraphics[width=\linewidth]{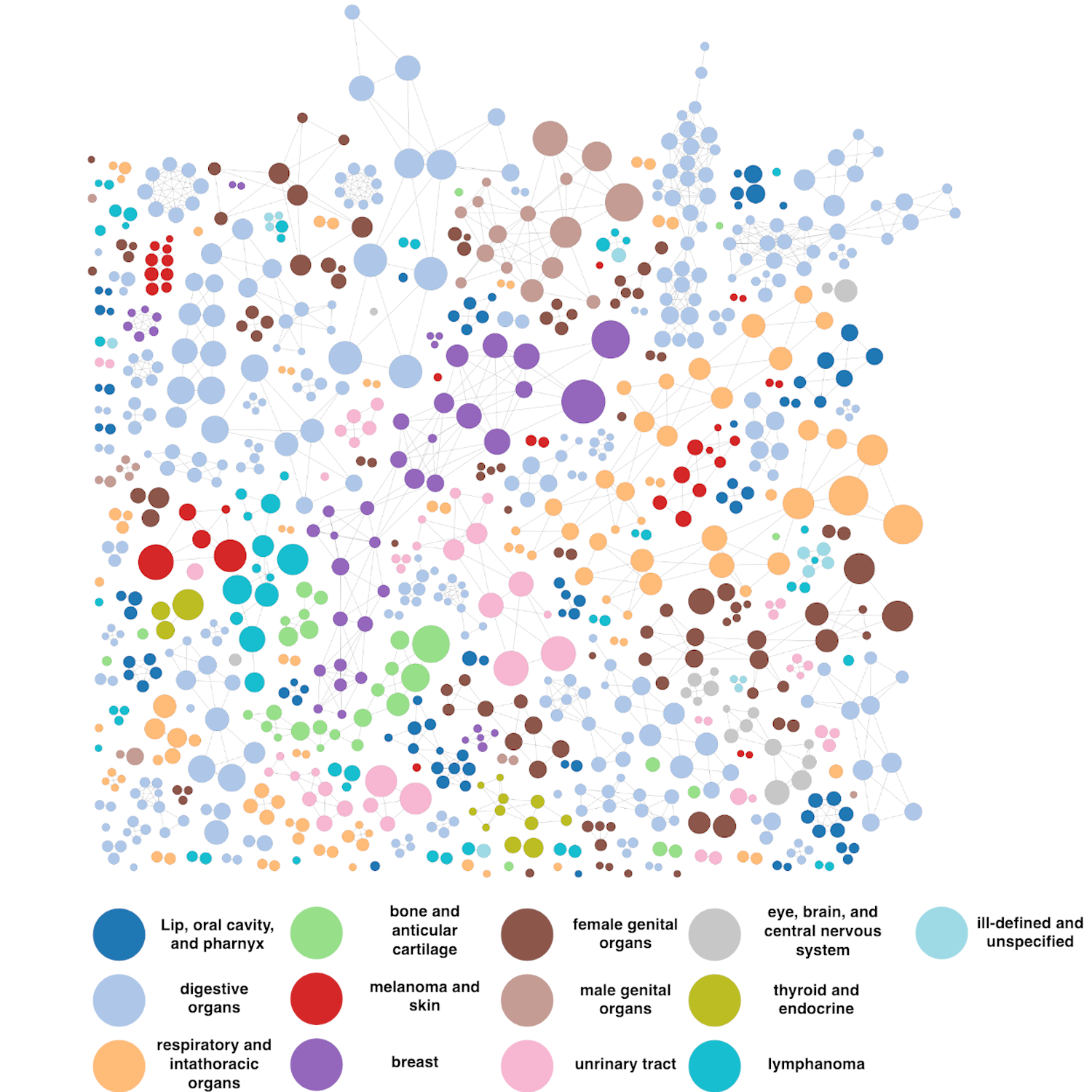}
		\caption{A visual representation of our mapper graph for the site task. 
		Colors designate the ground truth label, and the nodes are proportional to 
		the number of documents in the associated node. The purple subgraph at the center
		is associated with C50 (breast cancer), and the C61 (prostate cancer) cluster is 
		the tan nodes directly above the breast cancer subgraph.}
		\label{fig:site_mapper}
	\end{figure}

	Before being used in the information extraction task, each pathology report was 
	preprocessed according to the methodology described in~\cite{alawad2020automatic}.
 	Each of the remaining words in the cleaned documents were
	tokenized and padded to a uniform length of 3,000 tokens,
	and was used as input to the MTCNN~\cite{alawad2020automatic}. 
	\rev{The training set consisted of 236,519 reports, of which 10\% were set
	aside for a validation set. The testing set contained 78,856 reports and was
	trained to make simultaneous predictions on each of the cancer characteristics 
	(primary site (66 classes), grade (8 classes), histology (176 classes), laterality (7 classes), and behavior (4 classes)).}
	From our trained model,
	we then made multiple predictions on samples from the training set,
	allowing the dropout layer to vary between predictions. 
	\rev{The result of this step is a sampling of the model's feature space.
	Letting the dropout vary across multiple predictions on each document
	creates a representation of the model's learned feature space. It is these vector
	representations in the feature space of the pathology reports, which are used 
	for the final classification.} 
	Specifically, connections to the 
	subsequent layer were randomly activated with a probability of 0.5;
	this ensured differences in the 
	network connections, and consequently the confidence scores, 
	between subsequent predictions on
	each document. We then computed the average softmax score for each document and 
	took the class with the greatest score as the average predicted class for each document.
	We also computed the average softmax score of the ground truth class for each document.
	We used these two values, in conjunction with the ground truth label, and two-dimensional
	embedding~\cite{tenenbaum2000global} of the words in each document as inputs to create our mapper graph.
	These results are presented in~\cref{fig:site_mapper}.

%

	We then queried our mapper construction for each class in our
	dataset for those nodes in the graph that were classified at better than 97\%
	accuracy. These nodes contain a subset of the vocabulary that are associated
	with high predictive accuracy for a given label. However, not all the words in this 
	subset are indicative of any particular class. If we consider conjunctive words, they
	might tie together relevant words for a class but alone do not indicate any particular
	class label. Thus we want to infer the words with the highest probability of being 
	associated with a specific class; to approximate this probability, we use the distance to measure
	detailed in the previous section. For primary cancer sites, breast (C50) and prostate (C61),
	a table of the most prominent, i.e., closest to the high accuracy subset, is
	provided in~\cref{tab:site_words}.
		
	\begin{table}
	\caption{Label-specific keywords and associated density estimate
		inferred from our
		method computed over 2.3 million NCI cancer pathology reports for 
		primary cancer site C50 (breast) and C61 (prostate) with $\hat{m} = 0.25$.} \label{tab:site_words}
	\begin{tabular}{@{}ll|ll@{}}
		\toprule
		\multicolumn{2}{c}{C50}&\multicolumn{2}{c}{C61}\\
		Keyword & Density & Keyword & Density\\
		\midrule
absent  &  0.545855  &  cellular  &  0.384322  \\ 
htn  &  0.561171  &  firm  &  0.388699  \\ 
individual  &  0.568243  &  htn  &  0.390837  \\ 
obvious  &  0.580716  &  marked  &  0.392986  \\ 
marked  &  0.586395  &  examined  &  0.393910  \\ 
barrett  &  0.587200  &  deep  &  0.394185  \\ 
hyperplastic  &  0.587436  &  individual  &  0.397800  \\ 
kb  &  0.590752  &  hemorrhagic  &  0.400443  \\ 
hemorrhagic  &  0.596458  &  dense  &  0.400927  \\ 
herpes  &  0.598618  &  versus  &  0.403204  \\ 
gd  &  0.599345  &  iron  &  0.404446  \\ 
near  &  0.601415  &  hyperplastic  &  0.405272  \\ 
condensation  &  0.604456  &  numerous  &  0.407425  \\ 
follicles  &  0.606486  &  attempt  &  0.407799  \\ 
white  &  0.606775  &  degenerated  &  0.412005  \\ 
respectively  &  0.608007  &  specify  &  0.412277  \\ 
demonstrates  &  0.609365  &  present  &  0.413304  \\ 
specify  &  0.609693  &  absent  &  0.415616  \\ 
sparse  &  0.609972  &  obvious  &  0.415673  \\ 
two  &  0.610398  &  sparse  &  0.416259  \\ 
firm  &  0.612019  &  basophilia  &  0.416335  \\ 
cellular  &  0.612091  &  evaluated  &  0.416506  \\ 
normocellular  &  0.612507  &  blanchard  &  0.417760  \\ 
name  &  0.613017  &  minor  &  0.417820  \\ 
behavior  &  0.613755  &  laboratories  &  0.418889  \\ 
	\end{tabular}
	\end{table}	
	
\begin{figure}
		\centering
		\includegraphics[width=\columnwidth]{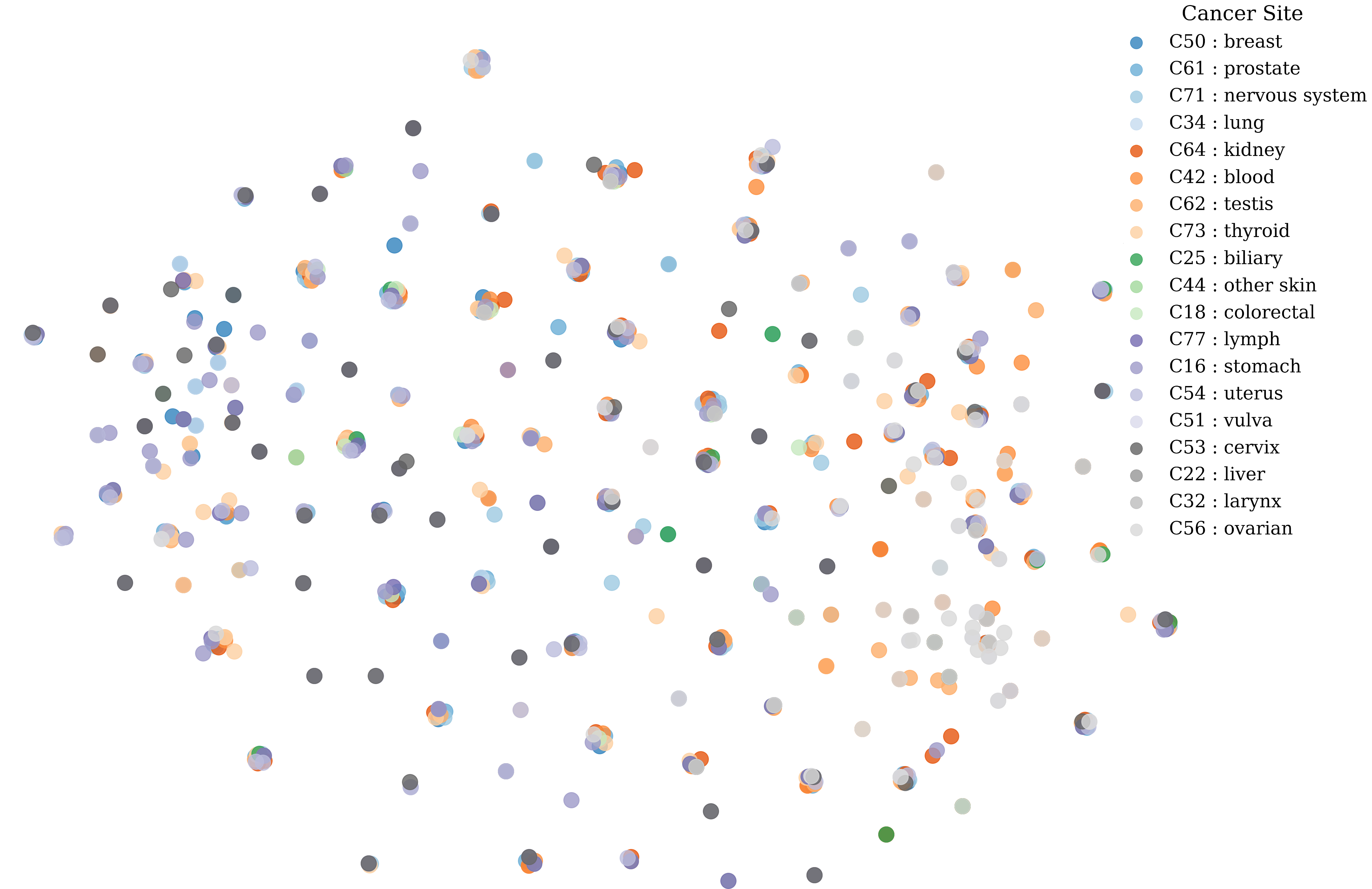}
		\caption{Two dimensional representation of the learned word embeddings
		from regions of high-predictive accuracy in our mapper graph. The clusters
		of words from multiple classes are common words indicating a cancer
		diagnosis, but are not specific as to a particular site; high-probability
		words are singletons typically found in regions with other words associated
		with the same class.}
		\label{fig:site_words_plot}
	\end{figure}

	The high-probability words in~\cref{tab:site_words} contain indicators of 
	disease that have been identified in the literature. For example, `pN' indicates prostate cancer
	with lymph node involvement ~\cite{amin2017eighth}, although the 
	last identifier in the code, typically an integer or \texttt{+} character,
	 has been stripped away in the data preprocessing, similarly for breast cancer
	 we see the term `ptnm'  which is the pathological tumor, node, metastasis (TNM) staging classification.
	 The specific code is omitted, due to differences between pathology reports, but we can 
	 see that this feature was identified as important across the entire collection 
	 of pathology reports.
	 Additionally, we note the word `htn', which
	is shorthand for `hypertension' and has been noted as an increased risk factor
	for both breast~\cite{han2017hypertension} and prostate~\cite{navin2017association} cancers.
	Additionally, `kb', in particular nuclear factor-kappa B (NF-$\kappa$B), is a biomarker for both 
	cancers as well~\cite{biswas2004nf,jin2008nuclear}. There are also words 
	contained in both columns of this table, such as `chromatin', `cellular',
	and `demonstrates', which are important to the structure of the reports, but are not indications 
	of a cancer diagnosis by themselves. However, as a majority of electronic pathology reports contain
	these words, it follows that they are identified as being part of the set of 
	words contained in documents with high-predictive accuracy. Clustering high-predictive words in
	two-dimensions across all primary sites in~\cref{fig:site_words_plot} reveals clusters for words
	indicative of a cancer diagnosis, but not site specific such as: cellular, hemorrhagic, and hyperplastic.
	Singletons are more specific to a particular site; prostate and lung are primary examples.
	
\subsection{20 Newsgroups Dataset}
	
	Due to the private nature of the information in the electronic health records, 
	we repeated our analysis on the publicly available 20 Newsgroup dataset~\cite{20News}. 
	Specifically, we created and trained a CNN with an architecture similar to~\cite{DBLP:journals/corr/Kim14f}
	on a subset of the 20 Newsgroups data. We used the `alt.atheism' \rev{(799 instances)}, `comp.graphics' \rev{(973 instances)}, 
	`sci.space' \rev{(987 instances)}, and `talk.religion.misc' \rev{(628 instances)} classes for our analysis.
	To preprocess step,
	we stripped out all of the headers and footers from the messages prior to tokenizing the 
	data, and removed non-alphanumeric characters and words shorter than three characters.
	The remaining data was split into train, test, and validation sets. 
	The training set was comprised of 3048 instances, of which 20\% were set aside 
	for a validation set, and the test set contained 339 samples. We trained our CNN to achieve 
	nearly 99\% accuracy on the training set using a cross-entropy loss function 
	and the Adam optimizer~\cite{kingma2014adam}.
	
\begin{figure}
		\centering
		\includegraphics[width=\linewidth]{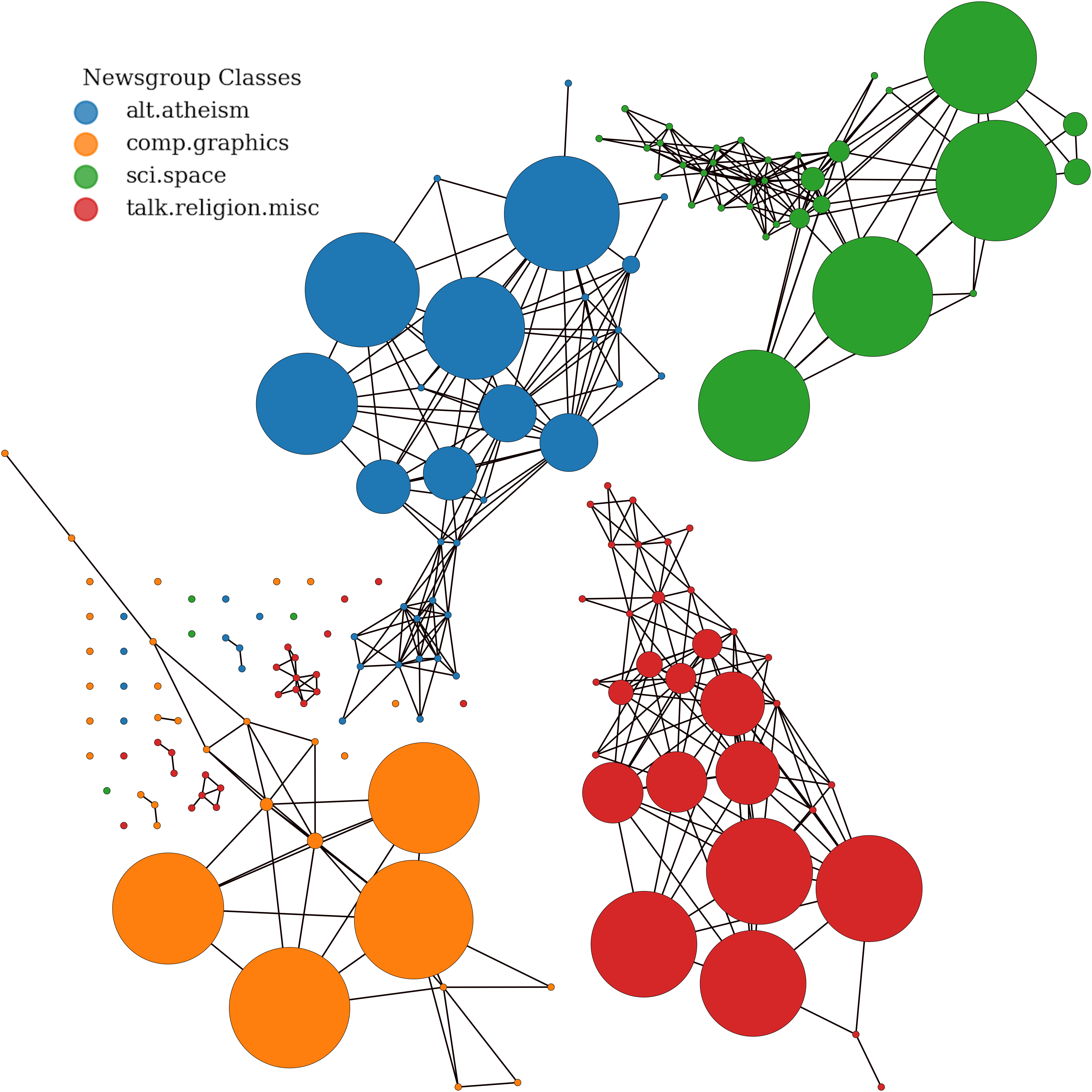}
		\caption{A visual representation of our mapper graph over the 20 Newsgroups subset.
		Colors designate the ground truth label, and the nodes are proportional to 
		the number of documents in the associated node. Note that not all subgraphs associated
		with a specific class label are connected. These smaller, disconnected
		regions contain documents that share similar text characteristics and are likely
		to be misclassified by our model.}
		\label{fig:ng_mapper}
	\end{figure}

	The results of our Mapper construction on this dataset is shown in~\cref{fig:ng_mapper}.
	For all classes, we observe large connected graphs 
	where the large nodes contain clusters of documents demonstrating high predictive accuracy
	from the CNN model. There are subgraphs from documents being from the `alt.atheism' or `talk.religion.misc' classes 
	that are disconnected from their associated primary graph. 
	These two classes have overlapping vocabularies, which potentially leads to
	misclassifications between these two classes, as opposed to the words from the `sci.space' class,
	which has few overlapping words with either of `alt.atheism' or `talk.religion.misc'. One interesting 
	overlap between these two classes and `sci.space' are words associated with planetary names from our solar system.
	For example, `Saturn' may refer to either the planet (`sci.space') or the god in Roman mythology, the context was
	inferred by the different parallel convolutional layers in the CNN model.
	
	\begin{figure}
		\centering
		\includegraphics[width=\linewidth]{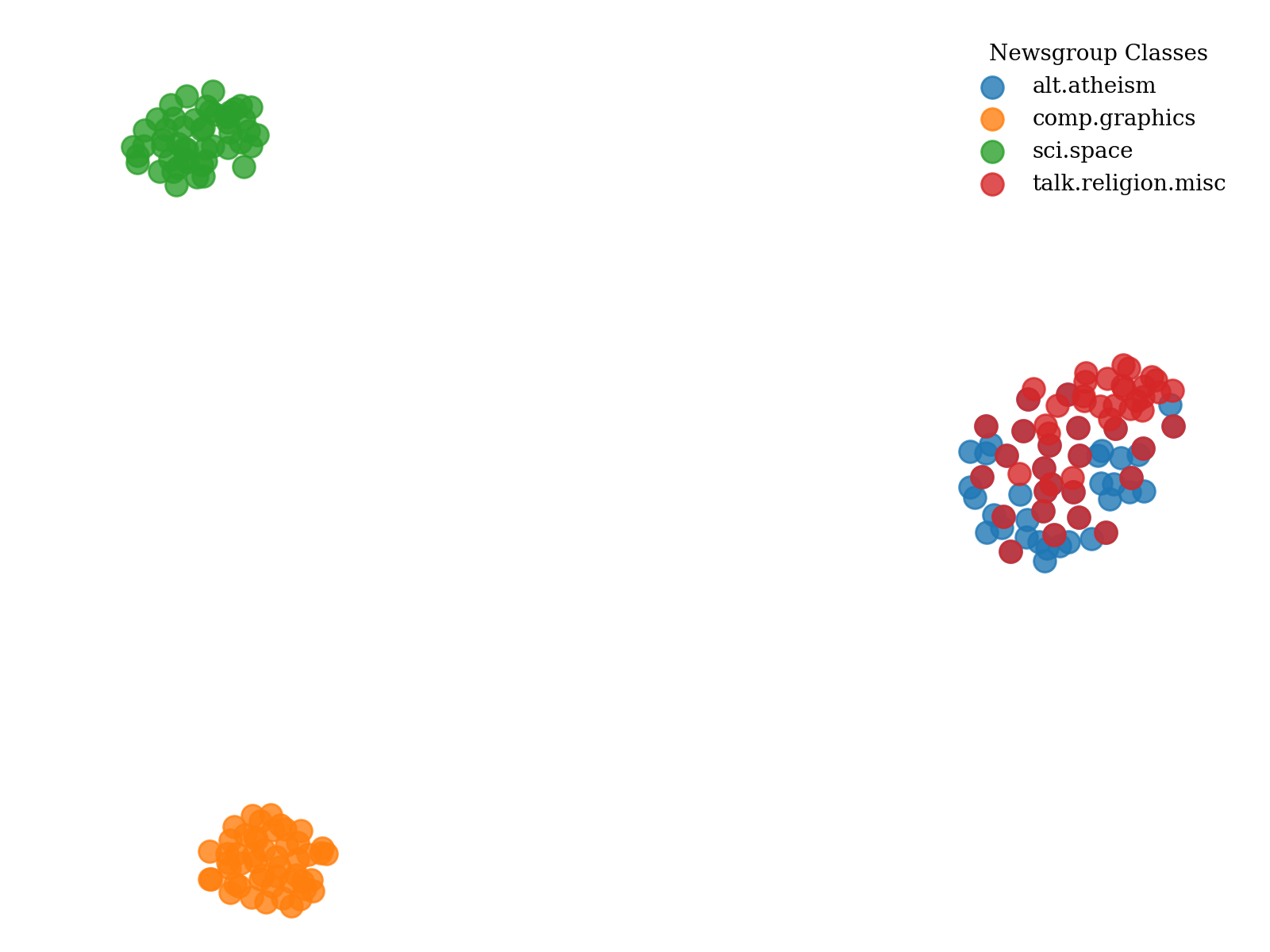}
		\caption{Two dimensional representation of the learned word embeddings for the 
		newsgroups dataset. These words are inferred
		from regions of high-predictive accuracy in our mapper graph. Observe that
		the classes are separated, except for the words associated with either
		alt.athesim or talk.religion.misc in the lower right-hand corner; 
		these latter two classes have significant overlap between their high-probability words.}
		\label{fig:ng_words}
	\end{figure}

	Computing the dtm, setting $m=0.1$, function for the different classes in the 20Newsgroups dataset, we 
	see the results from a t-SNE plot of the words associated with high-predictive accuracy in~\cref{fig:ng_words}.
	An interactive version of this plot is available in the git repository\footnote{https://code.ornl.gov/3t6/DL-interpretability}.
	We observe that
	the words associated with each class are well-separated, except for the 
	alt.atheism' or `talk.religion.misc' classes. This is expected behavior, as these classes
	have overlapping vocabulary, and discuss similar topics. Indeed, in~\cref{tab:ng_words01}
	we see the words `adultery' and `sinners' are
	strongly associated with both `alt.atheism` and `talk.religion.misc'.

	In~\crefrange{tab:ng_words005}{tab:ng_words025}, we study how different values of $\hat{m}$ in~\cref{eq:dtm}
	effect the results of our method, recalling that $\hat{m}$ is a smoothing parameter which 
	controls the amount of mass contained in balls centered at each point. An optimal value
	for $\hat{m}$ may be chosen in the case of a known target, but a distribution over words in
	the dataset is unknown. Furthermore, the behaviour of~\cref{eq:dtm} with respect to $\hat{m}$ is
	not monotonic~\cite{chazal2016rates}, which precludes typical approaches to model selection.
	The smaller values of $\hat{m}$ may introduce more variance to the results due to the
	global nature of the stability results between the empirical and theoretical quantile functions
	of the distribution in question.

	\begin{table}
	\centering
	\caption{Label-specific keywords and associated distance, with $\hat{m}=0.05$ approx.\,300 nearest neighbors,
		inferred from our
		method computed over 2{,}438 documents in the 20Newsgroups dataset from the four
		classes: alt.atheism, comp.graphics, sci.space, and talk.religion.misc.} \label{tab:ng_words005}
		  	\begin{adjustbox}{max width=\textwidth}
	\begin{tabular}{@{}ll|ll|ll|ll@{}}
		\toprule
		\multicolumn{2}{c}{alt.atheism}&\multicolumn{2}{c}{comp.graphics}&
			\multicolumn{2}{c}{sci.space}&\multicolumn{2}{c}{talk.religion.misc}\\
		Keyword & Distance & Keyword & Distance & Keyword & Distance & Keyword & Distance\\
		\midrule
mcl                  & 0.07152 & autodesk             & 0.06834 & moons                & 0.02539 & midst                & 0.08606 \\ 
healtasaturnwwcedu   & 0.07256 & pcs                  & 0.06951 & jupiters             & 0.03036 & wicked               & 0.08652 \\ 
timmbake             & 0.07324 & userdefined          & 0.07074 & jovian               & 0.03071 & servant              & 0.08712 \\ 
newnham              & 0.07691 & sunview              & 0.07246 & landers              & 0.03218 & preached             & 0.09280 \\ 
timmbakemclucsbedu   & 0.07727 & pixmap               & 0.07274 & rings                & 0.03221 & demons               & 0.09455 \\ 
nanci                & 0.07742 & histograms           & 0.07311 & flybys               & 0.03290 & souls                & 0.09469 \\ 
kempmpphoenixoulufi  & 0.07772 & printers             & 0.07335 & neptunes             & 0.03427 & lords                & 0.09650 \\ 
darksideosrheuoknoredu & 0.07798 & font                 & 0.07337 & neptune              & 0.03450 & fulfil               & 0.09675 \\ 
pihko                & 0.07809 & ultrix               & 0.07370 & nucleus              & 0.03484 & praise               & 0.09679 \\ 
ousrvroulufi         & 0.07832 & macintosh            & 0.07387 & lander               & 0.03579 & cupportalcom         & 0.09704 \\ 
	\end{tabular}
	\end{adjustbox}
	\end{table}	

	\begin{table}
	  \centering
	\caption{Label-specific keywords and associated distance, $\hat{m}=0.1$ approx.\,600 nearest neighbors,
		inferred from our
		method computed over 2{,}438 documents in the 20Newsgroups dataset from the four
		classes: alt.atheism, comp.graphics, sci.space, and talk.religion.misc. Words in red ink indicate changes from
		the previous table.} \label{tab:ng_words01}
		  	\begin{adjustbox}{max width=\textwidth}
	\begin{tabular}{@{}ll|ll|ll|ll@{}}
		\toprule
		\multicolumn{2}{c}{alt.atheism}&\multicolumn{2}{c}{comp.graphics}&
		\multicolumn{2}{c}{sci.space}&\multicolumn{2}{c}{talk.religion.misc}\\
		Keyword & Distance & Keyword & Distance & Keyword & Distance & Keyword & Distance\\
		\midrule
\new{unwilling}      &0.13562 & autodesk            &0.09517 & moons               &0.04566 & \new{sinners}             &0.12811 \\ 
\new{hatred}         &0.13575 & \new{halftone}            &0.09757 & jupiters            &0.04756 & \new{sins}                &0.12915 \\ 
\new{victims}        &0.13611 & sunview             &0.09804 & jovian              &0.04831 & \new{adultery}            &0.12942 \\ 
\new{practised}      &0.13616 & printers            &0.10066 & rings               &0.05272 & preached            &0.12989 \\ 
\new{oppression}     &0.13650 & pcs                 &0.10120 & landers             &0.05305 & \new{infant}              &0.13036 \\ 
\new{adultery}       &0.13655 & \new{xview}               &0.10139 & neptune             &0.05345 & \new{meat}                &0.13058 \\ 
\new{attacks}        &0.13711 & \new{fbm}                 &0.10313 & flybys              &0.05439 & \new{mormonism}           &0.13200 \\ 
\new{revenge}        &0.13790 & \new{nodes}               &0.10337 & neptunes            &0.05450 & wicked              &0.13335 \\ 
\new{imprisonment}   &0.13826 & \new{metafile}            &0.10436 & nucleus             &0.05564 & \new{victory}             &0.13419 \\ 
\new{sinners}        &0.14130 & \new{kernel}              &0.10486 & \new{crust}               &0.05759 & \new{selfrighteous}  &0.13444 \\ 
\new{protestant}     &0.14164 & font                &0.10530 & \new{volcanic}            &0.05804 & souls               &0.13606 \\ 

	\end{tabular}
	\end{adjustbox}
	\end{table}
	
	\begin{table}
	  \centering
	\caption{Label-specific keywords and associated distance, $\hat{m}=0.25$ approx.\,1600 nearest neighbors,
		inferred from our
		method computed over 2{,}438 documents in the 20Newsgroups dataset from the four
		classes: alt.atheism, comp.graphics, sci.space, and talk.religion.misc. Words in red ink indicate changes from
		the previous table.} \label{tab:ng_words025}
		  	\begin{adjustbox}{max width=\textwidth}
	\begin{tabular}{@{}ll|ll|ll|ll@{}}
		\toprule
		\multicolumn{2}{c}{alt.atheism}&\multicolumn{2}{c}{comp.graphics}&
		\multicolumn{2}{c}{sci.space}&\multicolumn{2}{c}{talk.religion.misc}\\
		Keyword & Distance & Keyword & Distance & Keyword & Distance & Keyword & Distance\\
		\midrule
unwilling           &0.20452 & autodesk            &0.16208 & \new{orbiters}            &0.12453 & selfrighteous       &0.19492 \\ 
practised           &0.20817 & printers            &0.16567 & landers             &0.12871 & meat                &0.19690 \\ 
adultery            &0.20937 & halftone            &0.16764 & jupiters            &0.13149 & adultery            &0.19924 \\ 
hatred              &0.21039 & nodes               &0.16776 & nucleus             &0.13460 & sinners             &0.20140 \\ 
imprisonment        &0.21110 & fbm                 &0.16980 & \new{ammonia}             &0.13558 & \new{inappropriate}       &0.20219 \\ 
victims             &0.21301 & sunview             &0.17079 & volcanic            &0.13576 & infant              &0.20333 \\ 
attacks             &0.21340 & \new{optional}            &0.17108 & flybys              &0.13587 & \new{hatred}              &0.20735 \\ 
\new{inappropriate}       &0.21410 & \new{standalone}          &0.17226 & \new{explored}            &0.13629 & sins                &0.20891 \\ 
\new{selfrighteous}       &0.21551 & \new{terminals}           &0.17545 & \new{polar}               &0.13712 & \new{insulting}           &0.20985 \\ 
protestant          &0.21644 & metafile            &0.17629 & \new{twin}                &0.13753 & \new{attacks}             &0.21010 \\ 

	\end{tabular}
	\end{adjustbox}
	\end{table}

	Comparing the results for `alt.atheism' change as the parameter $\hat{m}$ increases from 0.05 to 0.1,
	we notice a substantial change in the inferred features between the two runs. Most noticeably,
	in~\cref{tab:ng_words005} the words seemingly have no apparent connection to atheism. If we examine the
	documents individually, we see that the email addresses are associated with individuals who frequently
	post to the message board. Furthermore, some of the messages contain multiple responses, hence the addresses
	may appear multiple times within a single document and in multiple instances in the training set. In some
	sense, these words are most strongly correlated with the topic, since these individuals are frequent posters
	to the message board. However without looking back at the original data, these results are difficult
	to interpret. As the value of $\hat{m}$ is increased to 0.1, the most important words inferred through our method 
	for the `alt.atheism' class are words that require no further interpretation and are readily identified 
	with a discussion of religion and atheism. 
	
%
	
\subsubsection*{\rev{Unseen Examples}}
	
	\rev{Here we present results on a randomly chosen entry in the test split of the 20Newsgroups
	dataset and query our interpretability model. The text and associated results are shown
	in~\cref{tab:unseen}. We observe the most important words identified by our method
	are relevant to the document
	classification, but also contains some generic terms as well. While these common terms are not 
	indicative of one class or another, they are common throughout the dataset and could be 
	cleaned from the corpus during preprocessing.}
	
	\begin{table}
		\fbox{%
    \parbox{\linewidth - 2\fboxsep}{%
		\rev{`Hi there,\escape{n}\escape{n} I am here looking for some help. My friend is a interior decor designer. 
		He is from Thailand. He is\escape{n} trying to find some graphics software on PC. Any suggestion on which\escape{n} 
		software to buy, where to buy and how much it costs ? He likes the most\escape{n} sophisticated\escape{n} software
		(the more features it has, the better)'}
    	}%
	}\vspace{1em}
	\begin{tabular}{@{}ll@{}}
		\toprule
		\multicolumn{2}{c}{Keywords and dtm score}\\
		\midrule
		graphics & 0.1031 \\
		help & 0.1077 \\
		hi & 0.1132 \\
		pc & 0.1175 \\	
		features & 0.1248 \\	
		software & 0.1307 \\
		costs & 0.1307 \\
		looking & 0.1375\\
		trying & 0.1472 \\
		\bottomrule
	\end{tabular}
	\caption{\rev{Dtm results on an unseen datapoint from the test set.
			The predicted class is comp.graphics with probability 0.9842. The relevant words to the prediction 
			and associated scores are in the associated table.}}\label{tab:unseen}
	\end{table}
	
\subsubsection*{\rev{Stability}}

	\rev{In~\cref{tab:stability} we present the results from computing the label-specific
	Lipschitz constants for both datasets from our method and LIME. In both cases
	considered here, we observe that our interpretability results are more
	stable than those provided by LIME and SHAP.}

	\begin{table}
	\caption{\rev{Label-specific Lipschitz constants
		for our proposed method and LIME interpretability
		method computed over 500 NCI cancer pathology reports for the site task
		and 20 Newsgroup dataset taken from the test set.}} \label{tab:stability}
	\begin{tabular}{@{}l|llllll@{}}
		\toprule
		& Our Method & LIME & SHAP \\
		\midrule
		NCI Pathology Reports & 0.0013 & 0.570 & 0.529 \\
		20newsgropups data & 0.246 & 0.561 & 0.902\\
		\bottomrule
	\end{tabular}
	\end{table}

	\rev{We compare the stability of our method as compared against LIME by computing
	the label-specific Lipschitz constant which is defined as}
	\begin{equation}\label{eq:lipschitz}
		L(X_i) = \argmax_{X\in \mathcal{N}(X_i;\epsilon)}\frac{ \|g(X_i) - g(X)\| }{ \|X_i - X\| },
	\end{equation}
	\rev{where $\mathcal{N}(X_i;\epsilon)$ is the neighborhood of radius $\epsilon$ centered at $X_i$
	that is homogeneous with respect to the class label of $X_i$, and $g(\cdot)$ is the output
	from an interpretability method. As the Lipschitz value $L(X_i)$ measures
	the rate of change about a point $X_i$, we only consider those points in the neighborhood
	$\mathcal{N}(X_i;\epsilon)$ with the same label as $X_i$, since we would expect to see 
	a greater rate of change between different inputs from different classes.

In this experiment, we selected 500 documents at random from test set of the NCI pathology report
dataset and 20Newsgroups dataset, and use them as input into LIME, SHAP and our proposed method. 
We took the output from each method, computed~\cref{eq:lipschitz}, and reported the maximum value.
As in the previous results of~\crefrange{tab:ng_words01}{tab:ng_words025}, we observe that 
the explanations are stable with respect to the inputs. }

\subsection{Computational Aspects}
	
	\rev{The different parts of our method have very different computational considerations. 
	After training the initial model, we make multiple predictions on each instance within the
	training set, and compute aggregate prediction statistics and feature vectors 
	on a GPU with the Tensorflow deep learning framework~\cite{tensorflow2015-whitepaper}.
	From these aggregate feature vectors and prediction statistics, we construct 
	the Mapper graph in three steps: 
	\begin{itemize}
		\item projecting the feature vectors to a lower-dimensional space (from $\R^{900}$ to $\R^{100}$), 
		\item covering this projection with overlapping hypercubes, and 
		\item clustering the data within these cubes. 
	\end{itemize}	
	We use the KeplerMapper library~\cite{KeplerMapper}
	in our work, with a custom C implementation (and python wrapper) for the Hausdorff distance computations required
	for the Mapper parameter. The computational complexity of this metric is polynomial in the number of data points,
	which may be decreased though parallel computations or algorithmic developments~\cite{knauer2011directed,taha2015efficient}.
	Identifying the relevant words in the Mapper graph requires a $k$-nearest neighbor search in the
	word embedding space. The number of neighbors in the search is inversely proportional to $\hat{m}$
	and depends on the cardinality of $\mathcal{V}$. In particular, our word embeddings are in 
	$\R^{300}$ and in the experiments presented here $k\in[200, 2500]$. For these searches we
	used the PyNNDescent library~\cite{dong2011efficient} which provides a fast algorithm for approximate
	nearest neighbor lookups in high dimensional space.
	}

\section{Discussion}\label{sect:discussion}

	Our interpretability method demonstrates that our MTCNN cancer phenotype prediction model
	has identified decision rules that contain clinically relevant information. We have
	inferred a multiset of features associated with high-predictive accuracy for each
	class in the dataset, and identified those words closest to these multisets from
	each word in the vocabulary. This technique may also be used to identify those words that 
	are indicative of regions of poor-predictive accuracy, which yields two pieces of
	information. Primarily, these features may indicate errors in the ground truth class
	labels, which neither benchmark nor our datasets are immune from~\cite{northcutt2021pervasive}.
	This method allows us to identify both systematic misclassifications by our model and common traits between
	misclassifications with the same erroneous labels.
	
	As our interpretability method yields important features for each class, as an extension
	of this work, we may investigate the effects of training a model on a reduced vocabulary.
	Specifically, the vocabulary for the MTCNN model presented herein contains approximately
	178,\,000 words, each of which is embedded in a 300-dimensional space that is learned at training time.
	One could infer the words associated with high-predictive accuracy for each class, and use the 
	collection of these sets as a reduced vocabulary for training subsequent models. One would 
	expect to see similar accuracy metrics, e.g.\ , precision and recall,
	as a model trained on the full vocabulary but enjoying a reduced time-to-solution.
	Additionally, the density estimates computed from our method could be used to estimate the
	mutual information between the vocabulary and labels to reduce the size of the vocabulary. 

	The translational aspect of our method could be enhanced through the inclusion
	of longer word combinations in the inferred features from the MTCNN model. This would aid
	in differentiating between important phrases with different interpretations in different contexts.
	Moreover, investigating longer windows of text would aid in interpreting the results from
	our method, and would align with the derived features from the three parallel convolutional
	filters in the model architecture, recall these windows look at patterns $3,4,$ and $5$ words long
	in each of the convolutional layers. Other tunings could be easily implemented, for example, one
	could look at those words with the highest self-attention score in an attention based
	architecture~\cite{GAO2019101726}.
	
	One potential limitation of our method is the ability to look at features, words in the cases presented here,
	within their original context. Many words exhibit polysemy, whereby a word can have multiple meanings
	depending on the original context. For example, the word `bass' might refer to: a fish, a musical instrument,
	or low-frequency output. Only through context can we disentangle the intended meaning. Continuing on 
	with this work, we will extend our method to consider not only single words, but different length combinations
	which will give context to the inferred words and engender increased trust in the underlying model predictions.

\subsection{Conclusions}
	
 	We have presented a novel interpretability method that is based on the topological and
	geometric features in a DL model's feature space. Our method does not assume any form
	of the learned decision rules and is faithful to the training data.  
    We have demonstrated Lipschitz
    stability of our method as measured across inputs with the same ground truth label. 
	Moreover, our method infers a low-dimensional structure from a complex and high-dimensional dataset,
	shedding light on the black-box nature of DL algorithms. We may further develop the idea of low-dimensional
	representations by considering manifold approximation ideas from TDA coupled with the ideas
	presented here. 

\section*{Acknowledgments}

The authors gratefully acknowledge Xiao-Cheng Wu of the Louisiana Tumor Registry for
curating and providing the data.

This manuscript has been authored by UT-Battelle, LLC under Contract No. DE-AC05-00OR22725 
	with the U.S. Department of Energy. The United States Government retains and the publisher, 
	by accepting the article for publication, acknowledges that the United States Government 
	retains a non-exclusive, paid-up, irrevocable, world-wide license to publish or reproduce 
	the published form of this manuscript, or allow others to do so, for United States 
	Government purposes. The Department of Energy will provide public access to these 
	results of federally sponsored research in accordance with the DOE Public 
	Access Plan (http://energy.gov/downloads/doe-public-access-plan).

This research used resources of the Oak Ridge Leadership Computing Facility, which is a DOE 
Office of Science User Facility supported under Contract DE-AC05-00OR22725.

The authors would like to acknowledge the contribution to this study from other staff in the participating central cancer registries. 
These registries are supported by the National Cancer Institute’s Surveillance, Epidemiology, and End Results (SEER) Program, the 
Centers for Disease Control and Prevention’s National Program of Cancer Registries (NPCR), and/or state agencies, universities, 
and cancer centers.  
The participating central cancer registries include the following:
\begin{itemize}
\item Louisiana Tumor Registry working under contract numbers SEER:\\HHSN261201800007I/HHSN26100002 and NPCR: NU58DP0063.
\end{itemize}

\section*{Declarations}

\subsection*{Funding}

This work was completed under contract numbers: DE-AC05-00OR22725
and AC02200200100000.

\subsection*{Availability of data and materials}
 The code used to create
	the figures with the 20 Newsgroups data and interactive versions 
	of the figures are in a git 
	repository\\ \rev{\texttt{https://github.com/aspannaus/DL-interpretability}.}
	
\subsection*{Competing interests}
The authors declare that they have no competing interests.


\bibliographystyle{siam}
\bibliography{tda_refs.bib} 

\begin{thebibliography}{10}

\bibitem{tensorflow2015-whitepaper}
{\sc M.~Abadi, A.~Agarwal, P.~Barham, E.~Brevdo, Z.~Chen, C.~Citro, G.~S.
  Corrado, A.~Davis, J.~Dean, M.~Devin, S.~Ghemawat, I.~Goodfellow, A.~Harp,
  G.~Irving, M.~Isard, Y.~Jia, R.~Jozefowicz, L.~Kaiser, M.~Kudlur,
  J.~Levenberg, D.~Man\'{e}, R.~Monga, S.~Moore, D.~Murray, C.~Olah,
  M.~Schuster, J.~Shlens, B.~Steiner, I.~Sutskever, K.~Talwar, P.~Tucker,
  V.~Vanhoucke, V.~Vasudevan, F.~Vi\'{e}gas, O.~Vinyals, P.~Warden,
  M.~Wattenberg, M.~Wicke, Y.~Yu, and X.~Zheng}, {\em {TensorFlow}: Large-scale
  machine learning on heterogeneous systems}, 2015.
\newblock Software available from tensorflow.org.

\bibitem{agarwal2021neural}
{\sc R.~Agarwal, L.~Melnick, N.~Frosst, X.~Zhang, B.~Lengerich, R.~Caruana, and
  G.~E. Hinton}, {\em Neural additive models: Interpretable machine learning
  with neural nets}, Advances in neural information processing systems, 34
  (2021), pp.~4699--4711.

\bibitem{alawad2020automatic}
{\sc M.~Alawad, S.~Gao, J.~X. Qiu, H.~J. Yoon, J.~Blair~Christian,
  L.~Penberthy, B.~Mumphrey, X.-C. Wu, L.~Coyle, and G.~Tourassi}, {\em
  Automatic extraction of cancer registry reportable information from free-text
  pathology reports using multitask convolutional neural networks}, Journal of
  the American Medical Informatics Association, 27 (2020), pp.~89--98.

\bibitem{alvarez2018robustness}
{\sc D.~Alvarez-Melis and T.~S. Jaakkola}, {\em On the robustness of
  interpretability methods}, arXiv preprint arXiv:1806.08049,  (2018).

\bibitem{amin2017eighth}
{\sc M.~B. Amin, F.~L. Greene, S.~B. Edge, C.~C. Compton, J.~E. Gershenwald,
  R.~K. Brookland, L.~Meyer, D.~M. Gress, D.~R. Byrd, and D.~P. Winchester},
  {\em The eighth edition ajcc cancer staging manual: continuing to build a
  bridge from a population-based to a more ``personalized'' approach to cancer
  staging}, CA: a cancer journal for clinicians, 67 (2017), pp.~93--99.

\bibitem{begoli2019need}
{\sc E.~Begoli, T.~Bhattacharya, and D.~Kusnezov}, {\em The need for
  uncertainty quantification in machine-assisted medical decision making},
  Nature Machine Intelligence, 1 (2019), pp.~20--23.

\bibitem{Biau2011}
{\sc G.~Biau, F.~Chazal, D.~Cohen-Steiner, L.~Devroye, and C.~Rodr{\'\i}guez},
  {\em {A weighted k-nearest neighbor density estimate for geometric
  inference}}, Electronic Journal of Statistics, 5 (2011), pp.~204 -- 237.

\bibitem{biswas2004nf}
{\sc D.~K. Biswas, Q.~Shi, S.~Baily, I.~Strickland, S.~Ghosh, A.~B. Pardee, and
  J.~D. Iglehart}, {\em Nf-$\kappa$b activation in human breast cancer
  specimens and its role in cell proliferation and apoptosis}, Proceedings of
  the National Academy of Sciences, 101 (2004), pp.~10137--10142.

\bibitem{breiman2017classification}
{\sc L.~Breiman}, {\em Classification and regression trees}, Routledge, 2017.

\bibitem{carlsson2009topology}
{\sc G.~Carlsson}, {\em Topology and data}, Bulletin of the American
  Mathematical Society, 46 (2009), pp.~255--308.

\bibitem{carlsson2020fibers}
{\sc L.~S. Carlsson, M.~Vejdemo-Johansson, G.~Carlsson, and P.~G. J{\"o}nsson},
  {\em Fibers of failure: Classifying errors in predictive processes},
  Algorithms, 13 (2020), p.~150.

\bibitem{carriere2018statistical}
{\sc M.~Carriere, B.~Michel, and S.~Oudot}, {\em Statistical analysis and
  parameter selection for mapper}, The Journal of Machine Learning Research, 19
  (2018), pp.~478--516.

\bibitem{Chazal2011}
{\sc F.~Chazal, D.~Cohen-Steiner, and Q.~M{\'e}rigot}, {\em Geometric inference
  for probability measures}, Foundations of Computational Mathematics, 11
  (2011), pp.~733--751.

\bibitem{chazal2016rates}
{\sc F.~Chazal, P.~Massart, and B.~Michel}, {\em Rates of convergence for
  robust geometric inference}, Electronic journal of statistics, 10 (2016),
  pp.~2243--2286.

\bibitem{dong2011efficient}
{\sc W.~Dong, C.~Moses, and K.~Li}, {\em Efficient k-nearest neighbor graph
  construction for generic similarity measures}, in Proceedings of the 20th
  international conference on World wide web, 2011, pp.~577--586.

\bibitem{GAO2019101726}
{\sc S.~Gao, J.~X. Qiu, M.~Alawad, J.~D. Hinkle, N.~Schaefferkoetter, H.-J.
  Yoon, B.~Christian, P.~A. Fearn, L.~Penberthy, X.-C. Wu, L.~Coyle,
  G.~Tourassi, and A.~Ramanathan}, {\em Classifying cancer pathology reports
  with hierarchical self-attention networks}, Artificial Intelligence in
  Medicine, 101 (2019), p.~101726.

\bibitem{guo2017calibration}
{\sc C.~Guo, G.~Pleiss, Y.~Sun, and K.~Q. Weinberger}, {\em On calibration of
  modern neural networks}, in International conference on machine learning,
  PMLR, 2017, pp.~1321--1330.

\bibitem{han2017hypertension}
{\sc H.~Han, W.~Guo, W.~Shi, Y.~Yu, Y.~Zhang, X.~Ye, and J.~He}, {\em
  Hypertension and breast cancer risk: a systematic review and meta-analysis},
  Scientific reports, 7 (2017), pp.~1--9.

\bibitem{hatcher2002algebraic}
{\sc A.~Hatcher}, {\em Algebraic Topology}, Cambridge University Press,
  Cambridge, UK, 2002.

\bibitem{jin2008nuclear}
{\sc R.~J. Jin, Y.~Lho, L.~Connelly, Y.~Wang, X.~Yu, L.~Saint~Jean, T.~C. Case,
  K.~Ellwood-Yen, C.~L. Sawyers, N.~A. Bhowmick, et~al.}, {\em The nuclear
  factor-$\kappa$b pathway controls the progression of prostate cancer to
  androgen-independent growth}, Cancer research, 68 (2008), pp.~6762--6769.

\bibitem{DBLP:journals/corr/Kim14f}
{\sc Y.~Kim}, {\em Convolutional neural networks for sentence classification},
  CoRR, abs/1408.5882 (2014).

\bibitem{kingma2014adam}
{\sc D.~P. Kingma and J.~Ba}, {\em Adam: A method for stochastic optimization},
  arXiv preprint arXiv:1412.6980,  (2014).

\bibitem{knauer2011directed}
{\sc C.~Knauer, M.~L{\"o}ffler, M.~Scherfenberg, and T.~Wolle}, {\em The
  directed hausdorff distance between imprecise point sets}, Theoretical
  Computer Science, 412 (2011), pp.~4173--4186.

\bibitem{20News}
{\sc K.~Lang}, {\em 20 newsgroups}.

\bibitem{laugel2019dangers}
{\sc T.~Laugel, M.-J. Lesot, C.~Marsala, X.~Renard, and M.~Detyniecki}, {\em
  The dangers of post-hoc interpretability: Unjustified counterfactual
  explanations}, arXiv preprint arXiv:1907.09294,  (2019).

\bibitem{lum2013extracting}
{\sc P.~Y. Lum, G.~Singh, A.~Lehman, T.~Ishkanov, M.~Vejdemo-Johansson,
  M.~Alagappan, J.~Carlsson, and G.~Carlsson}, {\em Extracting insights from
  the shape of complex data using topology}, Scientific reports, 3 (2013),
  pp.~1--8.

\bibitem{lundberg2017unified}
{\sc S.~M. Lundberg and S.-I. Lee}, {\em A unified approach to interpreting
  model predictions}, Advances in neural information processing systems, 30
  (2017).

\bibitem{molnar2020interpretable}
{\sc C.~Molnar}, {\em Interpretable machine learning}, Lulu. com, 2020.

\bibitem{pml2Book}
{\sc K.~P. Murphy}, {\em Probabilistic Machine Learning: Advanced Topics}, MIT
  Press, 2023.

\bibitem{navin2017association}
{\sc S.~Navin and V.~Ioffe}, {\em The association between hypertension and
  prostate cancer}, Reviews in urology, 19 (2017), p.~113.

\bibitem{nicolau2011topology}
{\sc M.~Nicolau, A.~J. Levine, and G.~Carlsson}, {\em Topology based data
  analysis identifies a subgroup of breast cancers with a unique mutational
  profile and excellent survival}, Proceedings of the National Academy of
  Sciences, 108 (2011), pp.~7265--7270.

\bibitem{nielson2015topological}
{\sc J.~L. Nielson, J.~Paquette, A.~W. Liu, C.~F. Guandique, C.~A. Tovar,
  T.~Inoue, K.-A. Irvine, J.~C. Gensel, J.~Kloke, T.~C. Petrossian, et~al.},
  {\em Topological data analysis for discovery in preclinical spinal cord
  injury and traumatic brain injury}, Nature communications, 6 (2015),
  pp.~1--12.

\bibitem{northcutt2021pervasive}
{\sc C.~G. Northcutt, A.~Athalye, and J.~Mueller}, {\em Pervasive label errors
  in test sets destabilize machine learning benchmarks}, arXiv preprint
  arXiv:2103.14749,  (2021).

\bibitem{rathore2021topoact}
{\sc A.~Rathore, N.~Chalapathi, S.~Palande, and B.~Wang}, {\em Topoact:
  Visually exploring the shape of activations in deep learning}, in Computer
  Graphics Forum, vol.~40, Wiley Online Library, 2021, pp.~382--397.

\bibitem{rathore2023topobert}
{\sc A.~Rathore, Y.~Zhou, V.~Srikumar, and B.~Wang}, {\em Topobert: Exploring
  the topology of fine-tuned word representations}, Information Visualization,
  22 (2023), pp.~186--208.

\bibitem{ribeiro2016should}
{\sc M.~T. Ribeiro, S.~Singh, and C.~Guestrin}, {\em " why should i trust you?"
  explaining the predictions of any classifier}, in Proceedings of the 22nd ACM
  SIGKDD international conference on knowledge discovery and data mining, 2016,
  pp.~1135--1144.

\bibitem{rudin2018please}
{\sc C.~Rudin}, {\em Please stop explaining black box models for high stakes
  decisions}, Stat, 1050 (2018), p.~26.

\bibitem{shrikumar2017learning}
{\sc A.~Shrikumar, P.~Greenside, and A.~Kundaje}, {\em Learning important
  features through propagating activation differences}, in International
  conference on machine learning, PMLR, 2017, pp.~3145--3153.

\bibitem{simonyan2013deep}
{\sc K.~Simonyan, A.~Vedaldi, and A.~Zisserman}, {\em Deep inside convolutional
  networks: Visualising image classification models and saliency maps}, arXiv
  preprint arXiv:1312.6034,  (2013).

\bibitem{singh2007topological}
{\sc G.~Singh, F.~M{\'e}moli, and G.~E. Carlsson}, {\em Topological methods for
  the analysis of high dimensional data sets and 3d object recognition.}, SPBG,
  91 (2007), p.~100.

\bibitem{slack2020fooling}
{\sc D.~Slack, S.~Hilgard, E.~Jia, S.~Singh, and H.~Lakkaraju}, {\em Fooling
  lime and shap: Adversarial attacks on post hoc explanation methods}, in
  Proceedings of the AAAI/ACM Conference on AI, Ethics, and Society, 2020,
  pp.~180--186.

\bibitem{ai4sci-doe}
{\sc R.~Stevens, V.~Taylor, J.~Nichols, A.~B. Maccabe, K.~Yelick, and
  D.~Brown}, {\em Ai for science: Report on the department of energy (doe) town
  halls on artificial intelligence (ai) for science}, tech. rep., Argonne
  National Lab., 2 2020.

\bibitem{taha2015efficient}
{\sc A.~A. Taha and A.~Hanbury}, {\em An efficient algorithm for calculating
  the exact hausdorff distance}, IEEE transactions on pattern analysis and
  machine intelligence, 37 (2015), pp.~2153--2163.

\bibitem{tenenbaum2000global}
{\sc J.~B. Tenenbaum, V.~d. Silva, and J.~C. Langford}, {\em A global geometric
  framework for nonlinear dimensionality reduction}, science, 290 (2000),
  pp.~2319--2323.

\bibitem{KeplerMapper}
{\sc H.~J. van Veen, N.~Saul, D.~Eargle, and S.~W. Mangham}, {\em {Kepler
  Mapper: A flexible Python implementation of the Mapper algorithm}}, Oct.
  2020.

\end{thebibliography}

\end{document}